\documentclass[sigconf]{acmart}
\usepackage{amsmath,amsfonts}

\usepackage{algorithmic}
\usepackage{graphicx}
\usepackage{textcomp}
\usepackage{xcolor}
\usepackage{booktabs}
\usepackage{graphicx}

\usepackage{soul}
\usepackage{url}
\usepackage{graphicx}
\usepackage{float}
\usepackage{amsthm}

\usepackage{makecell}
\usepackage[switch]{lineno}
\usepackage{algorithmic}
\usepackage{subfig}
\usepackage{bm}

\usepackage[utf8]{inputenc}

\newcommand{\eg}{\textit{e.g.}}
\newcommand{\etal}{\textit{et al.}}

\newcommand{\inputX}{\textbf{X}}
\newcommand{\stateS}{\bm{s}}
\newcommand{\embedZ}{\bm{z}}
\newcommand{\featureV}{\bm{v}}
\newcommand{\repH}{\bm{h}}

\AtBeginDocument{%
  \providecommand\BibTeX{{%
    \normalfont B\kern-0.5em{\scshape i\kern-0.25em b}\kern-0.8em\TeX}}}

\begin{document}

\title{Event2Graph: Event-driven Bipartite Graph for \\ Multivariate Time-series Anomaly Detection}

\author{Yuhang Wu, Mengting Gu, Lan Wang, Yusan Lin, Fei Wang, Hao Yang}
\affiliation{%
  \institution{Visa Research, Visa}
  \city{Palo Alto, CA}
  \country{United States}
}

\begin{abstract}
Modeling inter-dependencies between time-series is the key to achieve high performance in anomaly detection for multivariate time-series data. The de-facto solution to model the dependencies is to feed the data into a recurrent neural network (RNN). However, the fully connected network structure underneath the RNN (either GRU or LSTM) assumes a static and complete dependency graph between time-series, which may not hold in many real-world applications. To alleviate this assumption, we propose a dynamic bipartite graph structure to encode the inter-dependencies between time-series. 
More concretely, we model time series as one type of nodes, and the time series segments (regarded as event) as another type of nodes, where the edge between two types of nodes describe a temporal pattern 
occurred on a specific time series at a certain time. Based on this design, relations between time series can be explicitly modelled via dynamic connections to event nodes, and the multivariate time-series anomaly detection problem can be formulated as a self-supervised, edge stream prediction problem in dynamic graphs. 
We conducted extensive experiments to demonstrate the effectiveness of the design.

\end{abstract}


\maketitle

\section{Introduction}
Detecting anomalies in time-series data has been an important problem in the research community of data mining as well as the finance industry. In many circumstances, anomaly patterns in multiple time-series need to be taken into account together to disclose the full picture of the system. For example, before a 
financial crisis taking place, multiple macro and micro economic indicators can get aberrant in a sequential manner. It is important to analyze their transition and anomaly patterns all together rather than treat each individual signal separately. In 
another smaller scope example, to model a merchant default risk, acquirer banks needs to collect multiple key business indicators such as cash flow, asset, liabilities, etc. To comprehensively monitor the financial health of the merchant, anomalies in multiple key indicators need to form up a story-line to describe the default probability of the merchant. 

Previous works in multivariate anomaly detection mainly rely on recurrent neural networks (RNNs).  Malhotra \etal \cite{DBLP:journals/corr/MalhotraRAVAS16} and Hundman \etal \cite{DBLP:conf/kdd/HundmanCLCS18} employed LSTM models \cite{DBLP:journals/neco/HochreiterS97} to capture the temporal dependencies of multivariate signals and adopted prediction and reconstruction errors, respectively, to identify the anomaly patterns. Su \etal \cite{DBLP:conf/kdd/SuZNLSP19} improved the classical RNN model by modeling the probability distribution of time series via incorporating stochastic variables. To further model the correlations of time-series explicitly, Zhao \etal \cite{zhao2020multivariate} proposed a graph attention network to propagate information from different time series and aggregate the information together before feeding into a GRU \cite{DBLP:journals/corr/ChungGCB14}. However, because of the underlying RNN structure, previous works assume a static, complete dependency graph among time series. These approaches may not perform well under regime change of time series where the underlying inter-dependencies are different. Our work is built upon the recent wisdom of dynamic graph neural network. We allow the connectivity of the graph changes dynamically in different time stamps based on the patterns on the time series. We believe this adaptive design is more realistic and flexible in the real-world. 
Experiments also show that this design leads to improved performance in multivariate anomaly detection.

\begin{figure}[!t]
	\centering
	\includegraphics[width=8.7cm, height=4.2cm]{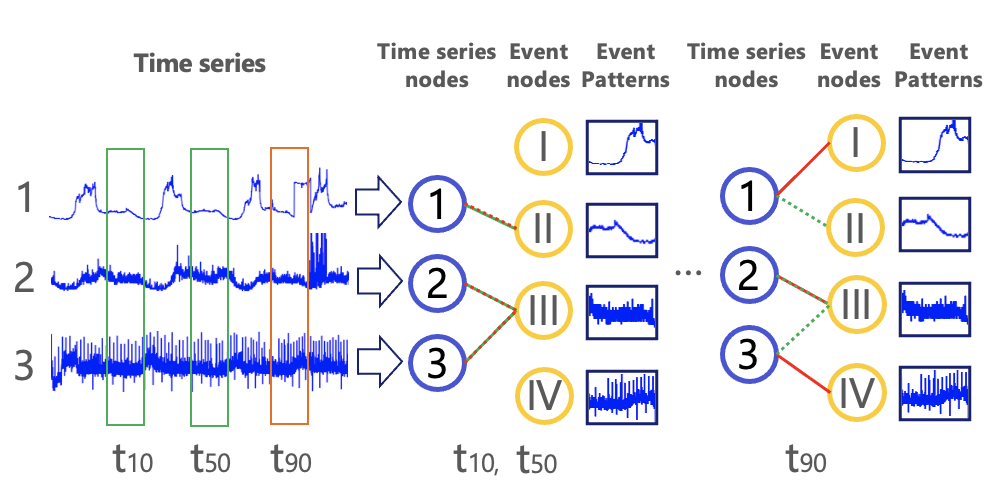}
	\caption{Overview of the event-driven bipartite graph structure. This toy example has six time-series nodes and three event nodes.   Time-step $\textbf{t}_{90}$ correspond to an anomaly pattern while other time stamps are normal. The relation between time sries nodes and event nodes are highlighted, green dashed line means the predicted relation while the red line means the actual relation.}
	\label{fig:1}
\end{figure}

\begin{figure*}[!t]
	\centering
	\includegraphics[width=15.7cm, height=3.4cm]{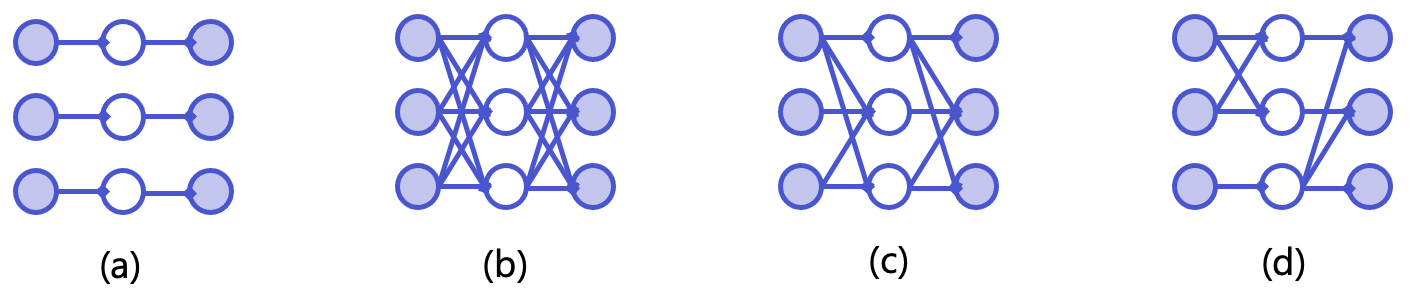}
	\caption{The inter-dependency map of multi-variate time series assumed by different time series forecasting models. (a) No inter-dependency assumed by uni-variate regression (\eg, ARIMA \cite{DBLP:conf/imc/ZhangGGR05}), (b) Static and complete inter-dependency assumed by RNN \cite{DBLP:journals/corr/MalhotraRAVAS16, DBLP:conf/kdd/HundmanCLCS18,DBLP:journals/neco/HochreiterS97,DBLP:conf/kdd/SuZNLSP19,zhao2020multivariate}, (c) static and partial inter-dependency assumed by graph structure learning \cite{deng2021graph}, (d) dynamic inter-dependency assumed by the proposed approach. }
	\label{fig:2}
\end{figure*}

In order to construct a dynamic graph on-the-fly, one important question is how to determine the connectivity of the graph in each timestamp. Previous works such as graph structural learning provide a feasible solution 
to constructing a static latent graph from time-series, but to the best of our knowledge, no previous work covers how to learn a dynamic and explainable graph for multivariate anomaly detection. Our work is inspired by the recent progress of evolutionary event graph \cite{DBLP:conf/aaai/Cheng00HZS20,DBLP:conf/wsdm/Hu0CYR21} where the nodes in the graph represent the time-sequences segments (events) and directed links represent the transition of the segments (events). Compare to previous works, this line of research naturally models the time-varying relations among time-series states via dynamic connections, and each state carries a physical meaning that is understandable by human. However, one major limitation of \cite{DBLP:conf/wsdm/Hu0CYR21} is that the event nodes employed in this work capture the information across all the time series. Assume there are $K$ segment patterns in each time series, and the number of time series is $D$. 
The model would need $K^D$ number of event nodes to represent the multivariate signals. This 
exponential number of event nodes strongly limits the information processing capability of the evolutionary state graph and therefore allows only a very small number of time series ($D$) to be analyzed in practice (four in the dataset used in the paper \cite{DBLP:conf/wsdm/Hu0CYR21}). 

To address the 
problem of exponential number of combinations, we disentangle the time-series nodes and the event nodes in our design, and model them as two types of nodes in a dynamic bipartite graph (as depicted in Fig. \ref{fig:1}). Each event node only 
represents a time segment on 
one individual time series, instead of integrating patterns across all time series. The undirectional connection between two types of nodes indicates event $e$ happens on the $d^{th}$ time series at time $t$. So the maximum number of edges in the graph is $O(KD)$, which is much smaller than $O(K^{2D})$. To further improve the efficiency and 
generalizability of the algorithm, we built upon the framework based on the recent advances in edge streams \cite{DBLP:conf/kdd/KumarZL19,xu2020inductive,DBLP:journals/corr/abs-2006-10637}, where connections between nodes are 
modeled as incoming attributed edges instead of constructing adjacency matrices. The complete system, with the name Event2Graph (Event-driven bipartite Graph), outperforms previous state-of-the-art on three public data-sets, and we summarize our main contributions as follows:

\noindent
(i) We propose a bipartite event-graph-based system to analyze the multivariate time series and model the interactions between time series and event segments via edge streams. This design significantly reduces the complexity of event encoding in multivariate time-series, and establishes state-of-the-art results in multivariate time-series anomaly detection.
\\
(ii) Events employed in the system is highly interpretable and easy for human to interact with. 
Through 
backtracing the historical event series or predicting future event series, the system can provide explainable root cause analysis and scenario-based forecasting.  \\

\section{Related work}
Most of the existing works in anomaly detection focus on univariate time series. In multivariate anomaly detection, based on how the methods model the dependencies among time series, they can be classified into two major categories, namely: ``static inter-dependencies"  approaches, and ``dynamic inter-dependencies" approaches. In this section, we introduce these two categories of methods in detail. We also summarize the recent progress in dynamic graph analysis, from where we derive the insight of the proposed solution. A schematic view is presented in \mbox{Fig. \ref{fig:2}} to introduce these methods.

\subsection{Static inter-dependency relation}
We argue that most of the existing multivariate anomaly detection models belong to this category of approach, where relations (\eg, correlation) between multivariate time series are fixed once learned, and the model assumes all the time series can influence each other (complete inter-dependency). LSTM-based framework 
has been widely employed in this category of work. For instance, Malhotra \etal  \cite{DBLP:journals/corr/MalhotraRAVAS16}  proposed a LSTM-based auto-encoder network to detect anomalies from multiple sensors, Su \etal \cite{DBLP:conf/kdd/SuZNLSP19} proposed a stochastic LSTM framework to model the data distribution of time series in a global perspective. As RNN was originally proposed to model the temporal dependencies between different timestamps, and all the dimensions of the input of RNN are used to describe a single concept (\eg word) all together, it was not tailored to model the inter-dependencies among variables by design. Recently, Zhao \etal \cite{zhao2020multivariate} proposed a graph attention network-based (GAT) approach \cite{DBLP:conf/iclr/VelickovicCCRLB18} where each time series is regarded as an individual node, and information are aggregated based on the underlying similarity of the signals. While this solution partially mitigates the problem by taking into account of dynamic pairwise similarities between time series, the design assumes a complete and static inter-dependency graph. Also, the processed information after GAT is simply aggregated all together and fed into a single GRU. Hence the module still suffers from similar problems as previous designs. 

\subsection{Dynamic inter-dependency relation}
A few recent works started to explore the partial inter-dependency relations among time series. Deng \etal \cite{deng2021graph} constructed a neighborhood graph on-the-fly based on sensor embeddings to describe the dependencies between different sensors. The sparsity level of this incomplete graph can be customized by users, but the connectivity of the graph is fixed once constructed. Hu \etal \cite{DBLP:conf/wsdm/Hu0CYR21} $\>$ proposed the first dynamic dependency design in multivariate anomaly detection. In this design, human interpretable time-series segments are used to create the nodes, and the transition of segments are explicitly modeled by dynamic graph neural network. This design allows the system to represent the time-varying relations among time series. However, one major limitation of the method is it uses a single event node to represent segments across all time series. This design brings the combination explosion problem in pattern representation (mentioned in Section I) and makes it hard to tackle a moderate scale problem.

\subsection{Dynamic Graph Neural Networks}
Our model is built upon the advances in dynamic graph neural networks. Most of the literatures in dynamic graph networks assume discrete-time graph dynamics where the graphs are represented as a sequence of snapshots \cite{DBLP:conf/ijcai/YuCACW17,DBLP:conf/kdd/YuCAZCW18,DBLP:conf/wsdm/SankarWGZY20}. Recent works start to explore more flexible design and assumes edges can appear at any time \cite{xu2020inductive,DBLP:conf/bigdataconf/NguyenLRAKK18,DBLP:conf/kdd/KumarZL19,DBLP:conf/iclr/TrivediFBZ19,DBLP:journals/corr/abs-2006-10637}. Our solution is derived from the state-of-the-art solution named Temporal Graph Networks (TGN) proposed by Rossi \etal \cite{DBLP:journals/corr/abs-2006-10637}. The model is built upon the temporal graph attention network (TGAT) \cite{xu2020inductive} but extended with an unique node-wise memory which attempt to model the temporal dependency on the node level instead of the system level as \cite{zhao2020multivariate}. This allow us to model the dynamic inter-dependency in our bi-partite event graph accurately, and provide us the flexibility to incorporate new nodes that have not seen in training.

\begin{table}[t]\centering
\caption{Notation.}

\begin{tabular}{cc}
\Xhline{\arrayrulewidth}
\inputX & Input time series \\
$D$  & Total number of time series sequences \\
$d$  & The $d^{th}$ time series sequence \\
$K$  & Total number of events \\
$\beta$ & Stride size of sliding window \\
$\tau$ & Length of sliding window \\
$\dot{t}$ & A single time step \\
$t$ & A time window \\
$\textbf{v}_e$  & Event node \\
$\textbf{v}_m$ & Time series sequence node \\
$\mathbb{G}^t_B$ & Bipartite event graph at time step $t$ \\
$A$ & Attributed edge between $\textbf{v}_e$ and $\textbf{v}_m$ \\
\hline
\Xhline{\arrayrulewidth}
\end{tabular}

\label{T20}
\end{table}

\section{Problem definition}

\textit{Problem definition: } Following \cite{DBLP:conf/wsdm/Hu0CYR21}, we assume the multivariate time-series composes multiple univariate time-series from the same entity, and the target problem can be defined as follows: A multivariate time-series is defined as $\inputX \in \mathbb{R}^{T \times d}$  where $T$ is the maximum number of time stamps in the time-series and $d$ is the dimension. The anomaly detection problem is to detect specific time stamps $\dot{t}^* \in \mathbb{O}$ in a multivariate time series where the time series behavior deviating from the normal patterns of the time series. Set $\mathbb{O}$ contains the timestamps that marked as anomaly by domain expert. In case $T$ is large, a common way to handle long time-series is to use a sliding window approach. Let $\tau$ denotes the length of the sliding window. We hereby formulate the problem as a binary classification problem with the objective to identify time windows with $\inputX^{[\dot{t}-\tau:\dot{t}] \times d} $ that contain anomaly time-stamps.

\textit{Background}: One solution of solving the multivariate time-series anomaly detection problem is to convert time series into a homogeneous graph \cite{DBLP:conf/aaai/Cheng00HZS20,DBLP:conf/wsdm/Hu0CYR21}, where $\mathbb{G}_e$ is defined on the representative patterns of a multivariate time series. The node of the graph $\textbf{v}_e$ corresponds to a human interpretable time series patterns $ \textbf{p} \in \mathbb{R}^{\tau \times d}$. The pattern should be representative enough so that the whole time series can be approximated by transitions of symbolic events (states) in the graph. The transitional relation between two sequential nodes (end with stamp $t$ and $t+1$) is defined as an edge in the graph. Many existing works (\eg, Bag of Patterns, Shaples, sequence clustering) can be used to distill representative patterns (event) from time series, but most of them only effective on univariate time series where $d=1$. 

\section{METHODOLOGY}

In this section, we present the proposed framework in detail. This section is organized as follows: In subsection 4.1,  we present the structure of our Event2Graph system and introduce how to use it for anomaly detection. Then, in subsection 4.2, we introduce how events nodes are created and added in the graph given a multivariate time series. Subsection 4.3 describes the network design to digest the information flow generated by Event2Graph. Finally, in Section 4.4, the optimization and inference methods are detailed.

\subsection{Bipartite Event Stream}
The core intuition behind the dynamic bipartite design is to decouple three key concepts in the traditional event graph, namely: where (which time series), when (at what time), and which (the event category). This practice helps us avoid the problem of exponential pattern combinations.

Specifically, the proposed algorithm is built upon a dynamic bipartite graph. We define it as a sequence of undirected bipartite event graph $\mathbb{G}^t_B = \{(\textbf{v}_m^t, \textbf{v}_e^t,  A(\textbf{v}_m^t, \textbf{v}_e^t))\}$. Here $t$ represents when the time window that the event graph is formulated. If the window size is set to be 1, then the time window $t$ is essentially the same as the time step $\dot{t}$. A time sequence index node $\textbf{v}_m^t$ indicates ``where" an event $\textbf{v}_e^t$ is happening. An attributed edge $A(\textbf{v}_m^t, \textbf{v}_e^t)$ that connect event node $\textbf{v}_e^t$ and sequence index node $\textbf{v}_m^t$ indicates an event $\textbf{v}_e^t$ happened on  time series $m$ at time window $t$. For simplicity, we also denote the edge as $A_{m,e}$ in the following sections. We employ an edge stream representation so an edge $A(\textbf{v}_m^t, \textbf{v}_e^t)$ is only constructed to represent the relation that actually existed. A major benefit of using edge stream representation over adjacency matrix is that it allows the graph structure to be more scalable and flexible, which provides us the generality of incorporating new events that have not appeared in training.

Based on this bipartite graph structure, we convert the multivariate anomaly detection problem into a self-supervised edge prediction problem in dynamic graph. Given the historical sequence of $\mathbb{G}_B^{(1:t)}$ and event nodes $\textbf{v}_e^{t+1}$ in time window $t+1$, we are predicting edges $\hat{A}(\textbf{v}_m^{t+1}, \textbf{v}_e^{t+1}))$ in $\mathbb{G}_B^{t+1}$. The anomaly is derived as a mismatching score from the predicted edge set $\hat{A}(\textbf{v}_m^{t+1}, \textbf{v}_e^{t+1}))$ and the observed edge set $A(\textbf{v}_m^{t+1}, \textbf{v}_e^{t+1}))$ with a read-out function $r(\cdot)$.

So the procedure of solving the multivariate anomaly detection problem becomes: 
\begin{enumerate}
    \item Given a normal sequence $\textbf{X}_{tr}$, identify representative patterns on each of the time sequence respectively in multivariate time series. 
    \item With identified representative patterns events, merge similar events across time-series to indicate affinity relation between time series.
    \item Build a sequence of bipartite event graph for multivariate time series (with a stride $\beta$) based on event matching. 
    \item Analyze the sequence of bi-partite event graphs and derive a model $\Psi$ to describe the intra- and inter-dependency relations.
    \item Given a testing sequence $\textbf{X}_{te}$, repeat (3), apply the model learned in (4) to predict $\hat{A}$ at time window $t^*$.
    \item Derive anomaly scores based on the predicted $\hat{A}$ and original time series with a proposed readout function $r(\cdot)$.
\end{enumerate}

\subsection{Event Node Detection And Matching}
We defined the event nodes on each time series separately and employed an unsupervised algorithm to identify the events. As we have no prior knowledge about what an anomaly pattern would look like in most real world applications, the system is designed to learn the representative patterns from observed normal data.

\textit{Representative segment detection}: We adopted Matrix Profile \cite{DBLP:conf/icdm/YehZUBDDSMK16}, a state-of-the-art unsupervised algorithm to identify representative patterns from time sequences. Matrix profile is able to identify the top-$K$ repeated patterns on time series with high accuracy and low computation time. We employed a recent implementation of matrix profile called SCRIMP++ \cite{DBLP:conf/icdm/0014YZKK18} on each of the time series in $\textbf{X}_{tr}$, and the algorithm yields a list of single dimensional representative patterns $\textbf{p}_{m,k}$ ($m \in \{1:D\}$, $k \in \{1:K\}$), each with size $\tau$.

\textit{Event generation}: many time series share similar representative patterns with each other. For example, many products sold in a supermarket may share similar seasonality behaviors with other products during a year, even though the actual amount may vary. Based on this observation, we argue that it is reasonable to merge similar segment patterns together across different time series to create the event nodes. Another advantage of merging detected segments is it helps to build the affinity relation cross time series. If two time series always share similar patterns, a spike happens in one time series that break the pair-wise correlation may be highly possible due to a true anomaly. To measure the similarity between time-series, we employed dynamic time wrapping \cite{DBLP:conf/kdd/BerndtC94}, which provides us with a robust distance matrix that is insensitive to small misalignments between two time series segments. After that, we employed the H-DBSCAN \cite{DBLP:journals/jossw/McInnesHA17} to cluster the patterns $\textbf{p}_{m,k}$ into clusters. We select the centroid of each cluster to be an event, which is a representative segment across all time-series within the cluster. Finally, we obtain an event set $\mathbb{E}_\tau$ which contains all events each with length $\tau$.

\begin{figure}[t]

    \includegraphics[width=9cm, height=3cm]{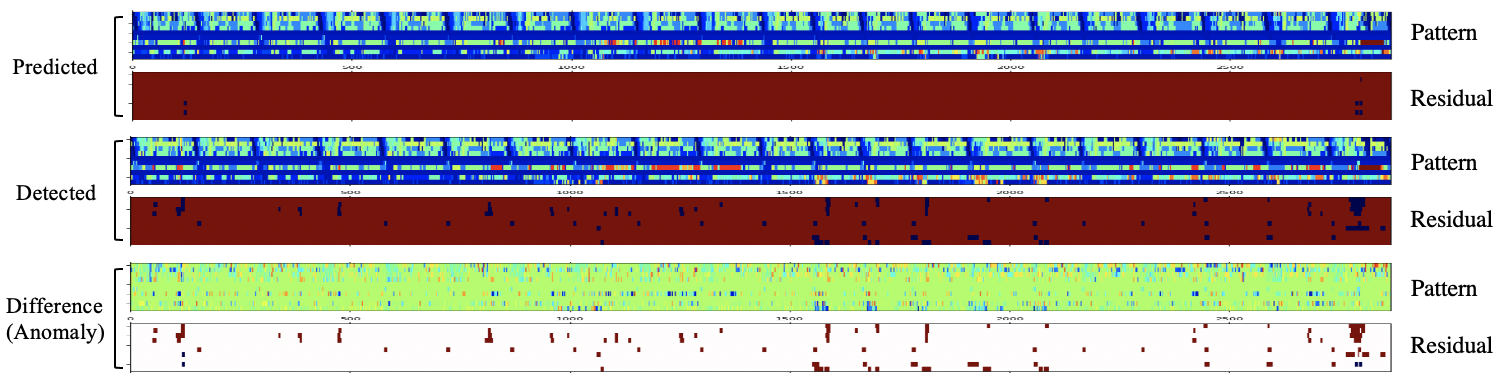} 
    \caption{Three group of event bars. Events are represented by colorful grids in the event bar. In each event bar, X axis corresponding to the time, each row corresponds a time-series. One event bar summarizes all the event happened on an individual time-series. For each group of event bar, two type event bars corresponding to the events generated by pattern matching and residual computing. The predicted event bars are generated by the proposed system via forecasting the next connected event node for each time-series. The anomaly score is generated based on the mismatch between the predicted events and the actual events with a carefully designed score function.}
    \label{eventbar}
\end{figure}

\begin{figure*}[t]
\centering
    \includegraphics[width=17cm, height=7cm]{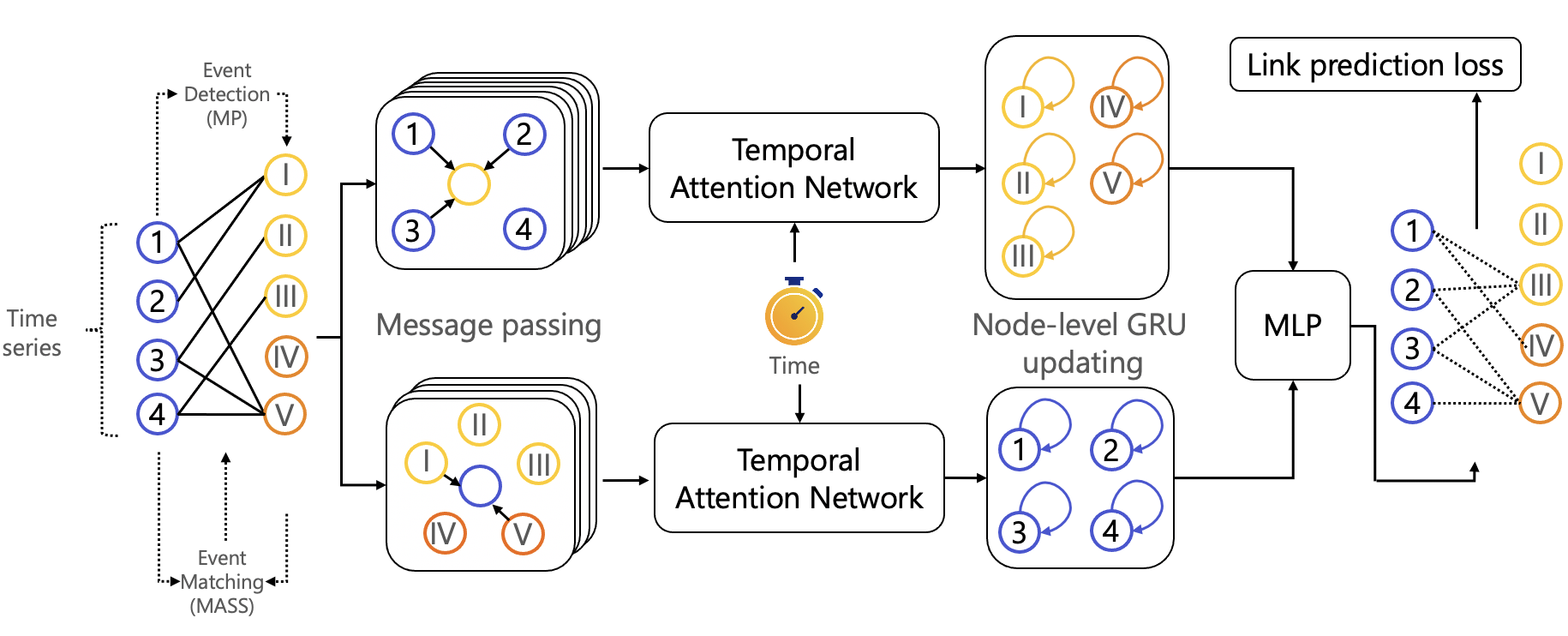} 
    \caption{Overview of the proposed approach. The time-series nodes are represented by blue circles, two types of event nodes (results from time series matching and residual computation) are represented by yellow and orange circle, respectively. }
    \label{pipeline}
\end{figure*}

\textit{Event matching}: after extracting $K$ events from time series segments, we match each event with the original time series to identify where and when each event is taking place. We employed the highly efficient C language-based Dynamic Time Wrapping algorithm \cite{FastestSimilaritySearch} to conduct the event matching, which provides a $(T-\tau) \backslash \beta \times D \times K$ similarity tensor $\mathcal{L}_\tau$ to indicate the similarity of each event to all the sliding window with size $\tau$ and stride $\beta$ in the input time sequence. 

\textit{Event nodes created by pattern matching:} 
for a single time stamp $t$, given $\mathcal{L}_\tau$ the event graph is created by selecting the best matched events for each time series. The event node is then connected with the corresponding time-series node $\textbf{v}_m^t$ in the event graph $\mathbb{G}^t_B$ with an attribute edge $A(\textbf{v}_e^t, \textbf{v}_m^t)$. By finding and linking the best matched event for each time series, we added $d$ number of event nodes into the graph. 

\textit{Event nodes created by residual error:} 
we noticed that even for the best matched time series, the pattern matching still left with small residual errors. We hereby define two general residual nodes which indicate whether the residual error in a time-series is larger than a threshold $\theta$. One residual event node denoted as $\textbf{v}_{e^+}^t$ indicates the residual error is larger than $\theta$, another residual event node denoted as $\textbf{v}_{e^-}^t$ indicates the residual error is equal or smaller than $\theta$. We learn $\theta$ in a data-driven manner with SPOT \cite{siffer2017anomaly} algorithm, where we employ the whole training data-set for initialization and testing for on-going adaptation. All time-series shared these two residual nodes as shown in Fig. 3.

 After generating a sequence of bipartite event graph $\mathbb{G}^{(1:t)}_B$ for time series $\textbf{X}_{tr}$, a model $\Psi$ is trained on $\mathbb{G}^{(1:t)}_B$ so that given a testing sequence $\textbf{X}_{te}$, it is able to predict the connectivity of an event graph in any specific time stamp $\dot{t}$ by observing historical sequence observed on $\textbf{X}_{te}$. In Fig. \ref{eventbar}, we visualize the events forecasted/detected on ten time-series. The difference between the top and the middle group result in the bottom group where we can easily identify the anomaly locations.

\subsection{Network Design With Node-wise Memory}
For each pair of $\textbf{v}_m^t$ and $\textbf{v}_e^t$ in graph $\mathbb{G}^{(t)}_B$, the node features of them are defined as $\featureV_m$ and $\featureV_e$, respectively. We also define $\bm{\epsilon}_{m,e}$ as the edge features between $\textbf{v}_m$ and $\textbf{v}_e$. In order to support the dynamic inter-dependency graph, we avoid explicitly defining a static adjacency matrix to describe the relation between nodes. Instead, we adopt a state-message framework to model the node interactions. For each node $\textbf{v}_m$ at time stamp $t$ (here we use time series node $m$ as an example; the same rule applies to the event node $e$), we define a state vector $\stateS_m(t)$ to represent its interaction history with other $\textbf{v}_e$ nodes before $t$ in a compressed format. By initiating $\stateS_m(0)$ as an all zero vector,  the interaction at time $t$ is encoded with a message vector $\bm{\varrho}_m(t)$: 
\begin{align}
    \bm{\varrho}_m(t) = [\bm{\epsilon}_{m,e}(t)|| \Delta t|| \stateS_m(t^-)|| \stateS_e(t^-)]
\end{align}
where $\Delta t$ is the time elapse between the previous time stamp $t^{-}$ and $t$, symbol || means concatenating operation. After aggregating all the messages from neighbors, the state vector of $\textbf{v}_m$ is updated as:
\begin{align}
   \stateS_m(t) = mem( agg\{\bm{\varrho}_m(t_1), ...,  \bm{\varrho}_m(t_b)\} ,\stateS_m(t^{-} ))
\end{align}
Here agg($\cdot$) is a general aggregation operation (support learnable parameters). For the sake of simplicity, we only compute the mean of the most recent messages in aggregation. We use mem($\cdot$) to represent a trainable update function (\eg, GRU).

Build upon the updated state vector $\stateS_m(t)$ and $\stateS_m(e)$, a time-aware node embedding can be generated at any time $t$ as following:
\begin{align}
    \embedZ_m(t) = \sum_{j \in n_m^k([0,t])}TGA(\stateS_m(t), \stateS_e(t), e_{m,e},\featureV_m(t), \featureV_e(t)) 
\end{align}
TGA represents the temporal graph attention module \cite{DBLP:journals/corr/abs-2006-10637}, where $L$ graph attention layers compute $m$'s embedding by aggregating information from its L-hop temporal neighbors.

\textit{Time-encoding}:
A finite dimensional mapping function is used to encode the time elapse between $t$ and $t_0$ as the functional time encoding: $\phi(t - t_0)$. The time encoding function allows us to encode time elapse with other graph features in an end-to-end manner. Specifically, the generic time encoding function used in \cite{rossi2020temporal} is employed, which is invariant to time rescaling, and can capture the periodical patterns in the data. 

\textit{Embedding update:} The TGA embedding module consists of a series of $L$ graph attention layers, which aggregates information $\tilde{\repH}_i^{(l)}(t)$ from each node's $L$-hop temporal neighborhood:
\begin{align}
    \repH_m^{(0)}(t) =& \stateS_m(t) + \featureV_m(t) \\
    \embedZ_m(t)  =& MLP^{(l)}(\repH_m^{(l-1)}(t)||\tilde{\repH}_m^{(l)}(t))  = \repH_m^{(l)}(t)
\end{align}
In each attention layer, a multi-head-attention is performed where a node attends to its neighboring nodes, generating key, query and values based on neighboring nodes' representation and the encoded time elapses. After temporal graph attention, an MLP (showing above) is used to integrate the reference node representation with the aggregated information:
\begin{align}
   \tilde{\repH}_m^{(l)}(t) =& MultiHeadAttention^{(l)}(\bm{q}^{(l)}(t),\bm{K}^{(l)}(t), \bm{V}^{(l)}(t))  \\
   \bm{q}^{(l)}(t) =& \repH_i^{(l-1)}(t) ||\phi(0)\\
   \bm{K}^{(l)}(t) =& \bm{V}^{(l)}(t)) \\
    =& [\repH_{e_1}^{(l-1)}(t) || \epsilon_{m,e_1}(t_1) ||\phi(t-t_1), ..., \\ &\repH_{{e}_N}^{(l-1)}(t) f|| \epsilon_{m,e_N}(t_N) ||\phi(t-t_N) ]
\end{align}

\subsection{Optimization and inference}
The aforementioned model is trained in a self-supervised fashion. As our goal is to predict the events that might happen in the time sequences in the next time step, which corresponds to the edges linking the time sequence nodes and event nodes. Therefore, we train our model with the edge prediction task. We use the cross entropy to capture the prediction loss, which is the standard loss function in edge prediction.

\textit{Event Forecasting Score:}  To convert the predicted event edges $\hat{A}$ into an anomaly score, for each time series node $\textbf{v}_{m,\tau}^{t}$ with window size $\tau$, the event  $\textbf{v}_{e,\tau}^{t}$ that has the highest probability connect to $\textbf{v}_{m,\tau}^{t}$ is retrieved and its pattern in the original signal space is denoted as $\textbf{s}^{t}_{e,\tau}$. We project the event $\textbf{s}^{t}_{e,\tau}$'s pattern back to its original signal space, we then compute anomaly score based on the dynamic time wrapping distance as follows: 
\begin{align} 
    \omega_{1,m}^t  = DTW (\textbf{X}^{t}_{m,\tau},  \textbf{s}^{t}_{e,\tau})
\end{align}

\textit{Residual Score:} For a positive residual event $\textbf{v}_{e^+}^t$ at time stamp $t$ where the forecasted results is not a positive residual $\hat{\textbf{v}}_{e^-}^t$, we calculate a changing point score to quantify the surprisal level as follows:
\begin{align} 
    \omega_{2,m}^t  = \psi_{NLG} (|| \textbf{X}^{t}_{m,\tau} - \textbf{X}^{t-\tau}_{m,\tau}||)
\end{align}
where $\psi_{NLG}$ is a standard function that maps a scalar into the negative log likelihood, which indicates the sparsity of this changing point in the training data. A frequent changing signal may results in small $\omega_{2,m}^t$ after the mapping. The function is learned in a data-driven manner based on the training data of time series $m$.

The final anomaly score at time stamp $t$ is calculated as one of the two following equations: 
\begin{align} 
   \omega^t_{max} = \max_{m}  (\omega_{1,m}^t \cdot \omega_{2,m}^t) \\
   \omega^t_{sum} = \sum_{m}  (\omega_{1,m}^t \cdot \omega_{2,m}^t)
\end{align}

The selection of the above equations is driven by the property of anomaly in a testing dataset. An empirical study shows that if very few (even a single) time-series determine the anomaly labels of the multivariate time series $t$, Eq. (14) is preferred. Otherwise, if the anomaly score of time $t$ is determined by multiple time series, in majority of cases, Eq. (15) would be a better choice. We evaluated both settings with comprehensive experiments. 

\section{Experiment}
We performed experiments on three datasets to demonstrate the effectiveness of our model on multivariate anomaly detection. We adopted three public datasets: SMAP (Soil Moisture Active Passive satellite) \cite{DBLP:conf/kdd/HundmanCLCS18}, MSL (Mars Science Laboratory rover) \cite{DBLP:conf/kdd/HundmanCLCS18}, and SMD (Server Machine Dataset) \cite{DBLP:conf/kdd/SuZNLSP19}. To compare with other SOTA models, we use the same set of metrics to evaluate our model, including precision, recall, and F1-score. 

We follow the evaluation protocol of \cite{su2019robust} which assumes anomalies occur continuously to form contiguous anomaly segments and an alert is classified as a correct catch if it is triggered within any subset of a ground truth anomaly segment.

\begin{table}\centering
	\caption{Dataset statistics and hyper-parameters}
\scalebox{1.0}{
\begin{tabular}{cccc}

\Xhline{\arrayrulewidth}
 \hline
                               & SMAP   & MSL  & SMD  \\
  \hline
Number of Time series nodes    &   25   &  55  & 38\\
Number of event nodes          &  130   & 70   & 64 \\
Time window length ($\tau$)    &  20    & 20   & 50 \\
Time window stride ($\beta$)   &  5     & 5    & 10 \\
\cline{1-3}
\hline
\Xhline{\arrayrulewidth}
\end{tabular}}
\label{Benchmark}
\end{table}

\subsection{Settings}
The proposed bipartite event graph contains $D+K$ number of nodes, where $D$ is the number of time series and $K$ is the number of events. The detailed number of nodes is shown in Table 2. The length ($\tau$) and stride ($\beta$) of the time window of each dataset are also displayed in Table 2. Our model can be easily extended to a multi-scale version by incorporating more event nodes with varieties of window sizes and strides. In this work, to keep things simple, we demonstrated our method on a single time-scale. For each time series, we set the maximum number of motifs detected to be three, and the minimum cluster size of H-DBSCAN to be three. For the temporal attention model, we set the number of multi-head to be 2. The GRU is employed to model the time encoding. The dimension of node embedding and message vector are both set to 64 respectively. 
Each model is trained after 10 epochs with the learning rate 0.0001. The node features of both time-series and event nodes are randomly initialized, and edge features are the one-hot embedding of the numerical ID of the nodes on both sides of the edge.

\begin{figure}[t]

    \includegraphics[width=8.5cm, height=3cm]{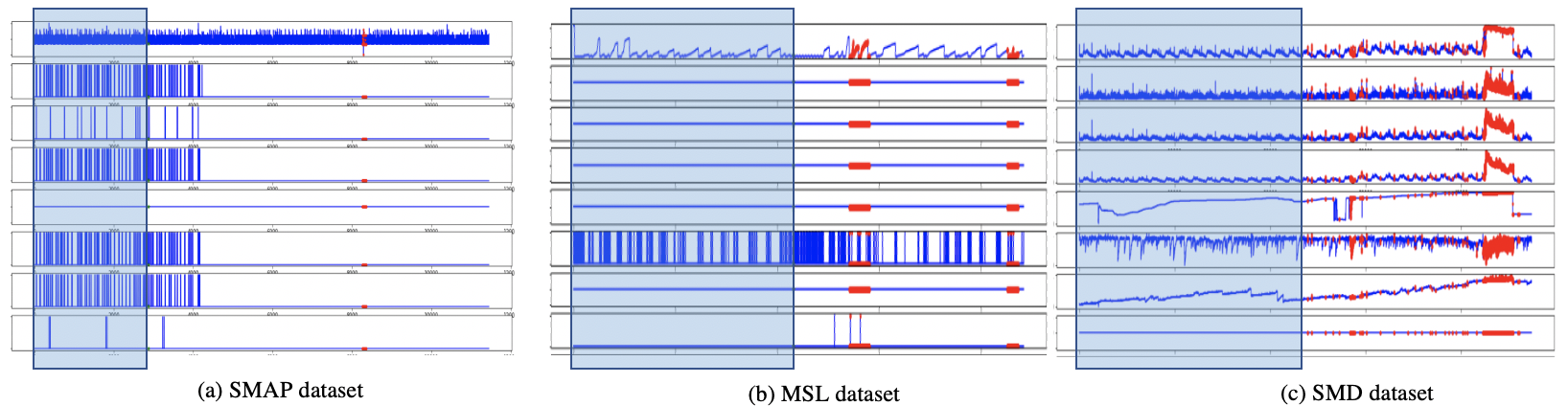} 
    \caption{The multivariate time-series anomaly on each of the three datasets: SMAP, MSL, and SMD. The training partition is highlighted in blue.}
    \label{eventbar}
\end{figure}

\subsection{Compare with state-of-the-art}
We compared our solution with multiple baselines on SMAP, SMD, and MSL datasets: DAGMM \cite{DBLP:conf/iclr/ZongSMCLCC18} - an autoencoder-based anomaly detection model without taking into account of temporal information; LSTM-VAE \cite{8279425}, LSTM-NDT \cite{DBLP:conf/kdd/HundmanCLCS18}, two state-of-the-art LSTM-based anomaly detection solutions; and the most recent stochastic VAE-based approaches (\eg, Omni-Anomaly\cite{su2019robust}) and graph attention-based method such as MTAD-GAT\cite{zhao2020multivariate}. We selected these baselines mainly because: (i) they are self-supervised algorithms that do not need any training labels (different from \cite{DBLP:conf/wsdm/Hu0CYR21}), (ii) they rely on a single scale of time-window (instead of multi-scale \cite{DBLP:conf/nips/ShenLK20}) so that the performances are directly comparable. 

The results are reported in Table \ref{Benchmark}. From the results we observed that the proposed Event2Graph achieves new state-of-the-art performance on SMAP dataset, ranked 2$^{nd}$ on SMD dataset, and perform very competitive on MSL dataset. In SMAP dataset, we observe that some of the data suffers from significant regime change during both training and testing, and our dynamic graph-based solution helps the algorithm adapt to the regime change faster than MTAD-GAT\cite{zhao2020multivariate} and Omni-Anomaly\cite{su2019robust}. We also observe that our algorithm significantly outperforms simple LSTM-based solution (LSTM-VAE and LSTM-NDT), which assumes a complete inter-dependency graph. They did not model the dynamic inter-dependency among time series, while our node-level model explicitly encode the temporal information along with the attention, which helps to reduce the false alarms in anomaly detection. Furthermore, since DAGMM assumes a completely static relationship between time series, the algorithm lacks of the capability to adapt to any temporal evolving pattern.

\begin{table}\centering
	\caption{Compare with current state-of-the-art approaches. The top three performance algorithms are highlighted.}
\scalebox{1.0}{
\begin{tabular}{ccccc}

\Xhline{\arrayrulewidth}
 \hline
Dataset         & Method &   Precision  & Recall & F1  \\
  \hline
  
               & DAGMM   &   59.51  &  88.82  & 70.94\\
               & LSTM-VAE &  79.22 &  70.75  & 78.42 \\
               & LSTM-NDT &  56.84   &  64.38  &  60.37 \\
 SMD           & OmniAnomaly  & 83.34 & 94.49   & \textbf{88.57}  \\
               & MTAD-GAT     & - & -  & -  \\
               & Event2Graph (max)  & 85.35   & 83.71 & \textbf{83.47}  \\
               & Event2Graph (sum) & 88.61 & 83.38  & \textbf{84.93}  \\

 \cline{1-5}
               & DAGMM   &  54.12   &  99.34  & 70.07 \\
               & LSTM-VAE & 52.57    &  95.46  & 67.80 \\
               & LSTM-NDT &  59.34   & 53.74   & 56.40 \\
 MSL           & OmniAnomaly   & 88.67    & 91.17 & \textbf{89.89} \\
               & MTAD-GAT    &  87.54    & 94.40 & \textbf{90.84}   \\
               & Event2Graph (max)  & 81.21  & 89.69  &  85.24  \\
               & Event2Graph (sum)  &  88.12  &  83.11 & \textbf{85.55}  \\
              
  \cline{1-5}       
                 & DAGMM   & 58.45    &  90.58  & 71.05 \\
                 & LSTM-VAE &  85.51   & 63.66  & 72.98 \\
                 & LSTM-NDT & 89.65    &  88.46  &  89.05 \\
  SMAP           & OmniAnomaly  & 74.16  & 97.76 & 84.34  \\
                 & MTAD-GAT    & 89.06   &  91.23 & \textbf{90.13}      \\
                 & Event2Graph (max)     & 90.48 & 93.21 & \textbf{91.82}  \\
                 & Event2Graph (sum)    &  86.54    &   94.33   & \textbf{90.27} \\

 \cline{1-5}

  \hline
\Xhline{\arrayrulewidth}
\end{tabular}}
\label{Benchmark}
\end{table}

We also observe that the summation-based aggregation (Eq. 14) generally outperforms max-based aggregation (Eq. 13). However, on SMAP dataset, the summation-based performs quite well. One reason is that the anomalies on SMAP dataset are mainly caused by a few time series, so the summation may bring extra noise if the algorithm takes into account all the time series together. However, the anomaly in SMD and MSL dataset may occur in a concurrent manner where the summation operation boosted the true positives that concurrently happened, while suppress false-alarms that happen individually on a specific time series. From now on, the default setting of Event2Graph would be with summation-base aggregation.

\subsection{Quantitative Ablation study}
The objective of this experiment is to provide detailed analysis over the effectiveness of each proposed module. The experiments are conducted on SMD dataset. By removing each of the critical module of our model, the performances are reported in \mbox{Table \ref{Ablation}}. 

\begin{table}\centering
	\caption{Ablation study on SMD dataset.}
\scalebox{1.0}{
\begin{tabular}{cccc}

\Xhline{\arrayrulewidth}
 \hline
Model         &   Precision  & Recall & F1          \\
  \hline
Event2Graph  &  88.61 & 83.38  & 84.93   \\
w/o TGA  &  84.81    & 82.94   & 82.51   \\
w/o  score $\omega_{1}$    &  82.83   &  80.29  &   79.45   \\
w/o  score $\omega_{2}$      &  81.51   & 71.76   &  73.81    \\
w/o TGA and w/o  $\omega_{1}$  &  80.56   &  81.09  &   79.47   \\
w/o  TGA and w/o  $\omega_{2}$      &  75.84    & 64.33   & 64.51   \\
$\psi_{NLG}$ only   &  85.89    & 75.22   & 78.48   \\
  \hline
\Xhline{\arrayrulewidth}
\end{tabular}}
\label{Ablation}
\end{table}

\subsubsection{Effectiveness of temporal graph attention}
Through replacing the temporal graph attention with a simple MLP module, the model's F1 score is reduced by 2.42\%. It demonstrated that the temporal attention plays an important role in aggregating information from neighborhood nodes. 

\subsubsection{Effectiveness of event forecasting score $\omega_{1}$:}
We remove the event matching score (Eq. 11) ) and just use a residual score (Eq. 11) for anomaly detection. We observe that the model performance drop by 5.48\%. The experiment indicates the event forecasting module in Event2Graph is essential for accurate anomaly detection. It helps the system understand the normality of time series, model the context of changing point, and work jointly with the residual score to achieve high model performance. From the experiment, we also observe that TGA plays a critical role in guarantee the quality of the forecasting score $\omega_{1}$. By removing TGA, the forecasting-based model performance (w/o  score $\omega_{2}$) degrade from 73.81\% F1 score to 64.51\%. 

\subsubsection{Effectiveness of residual score  $\omega_{2}$}

We remove the residual score (Eq. 12) and only use forecasting score to detect anomaly. The model suffers more than 10\% performance degradation. This results shows that modeling the residual score is essential to allow the system to identify the anomaly regions that may not well characterized by the event matching scores. Since this score mainly captures the abrupt change of values in time-series, we can infer that the anomaly pattern in SMD dataset are mainly due to the changing points. We also observe that the TGA may not contribute much to residual event modeling, a $\omega_{2}$ only model perform competitively well with or without TGA. This phenomenon tells us the residual event is something unexpected and an accurate changing point detection is better than learning simple, repeatable patterns.

\subsubsection{Effectiveness of log likelihood function $\psi_{NLG}$}
To demonstrate our assumption, we compare the results with a baseline that using a naive changing point detection score $\psi_{NLG}$, without adopting any forecasting or event modeling model. Surprisingly, the changing-point only model is able to achieve 78.48\% F1-score. Outperforms DAGMM, LSTM-VAE, as well as LSTM-NDT. This results further confirm that it is hard to learn any static pattern by using LSTM or autoencoder for anomaly detection on a challenge dataset, a dynamic model that is able to adapt to the changing of time-series patterns is preferred.

\subsection{Qualitative Ablation study}

\begin{figure*}[ht]%
\centering
\includegraphics[width=18cm, height=5cm]{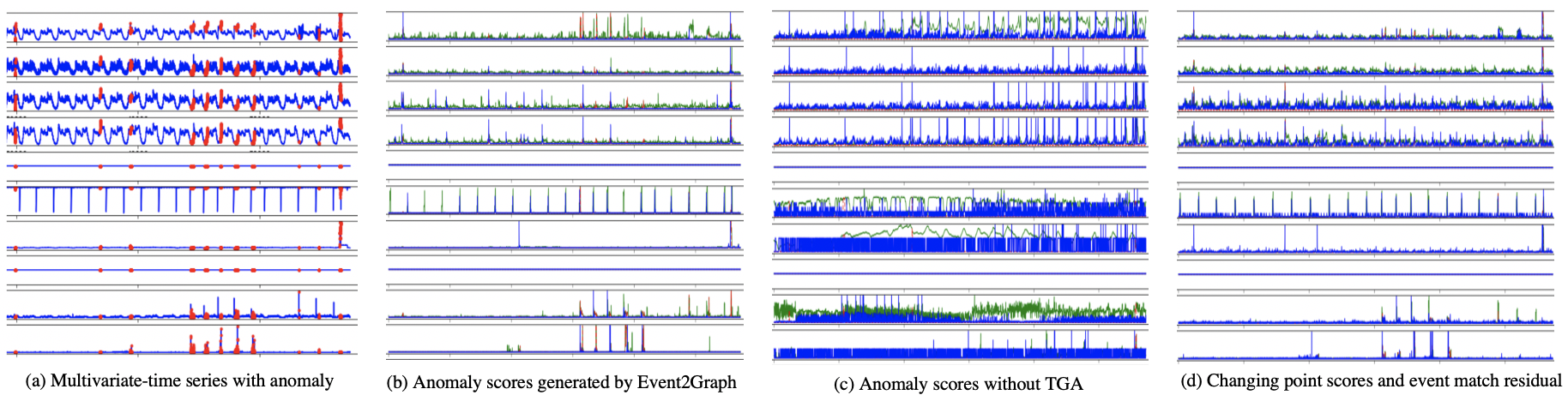}
\caption{Visualization of the event forecasting score $\omega_{1}$ and residual scores $\omega_{2}$ on a randomly selected machine in the SMD dataset. The top 10 time-series of this machine is visualized via (a), and the ground-truth anomaly in all sub-figures are highlighted with red. Subfigure (b,c) corresponding to the estimation of $\omega_{1}$ (green line) and $\omega_{2}$ (blue line) under different settings of Table 4. In (d), the change point score generated by $\psi_{NLG}$ is marked in blue, green line corresponding to the minimal residual error by matching all possible event nodes.}
\label{fourplot}
\end{figure*}

To better understand the behavior of our Event2Graph approach, we compared the propose approach with other variations via visualization. We present the two scores ($\omega_{1}$ and $\omega_{2}$) generated by the system on a randomly selected machine from SMD dataset. The original time series of this machine is shown in Fig. \ref{fourplot}(a). The $\omega_{1}$ and $\omega_{2}$ are visualized in Fig. \ref{fourplot}(b-d) under different experiment protocol. From the experiment, we observe that the temporal graph attention plays a very important role in terms of stabilizing the system. By removing this module, the whole system seems to overreact to small noises in the time series. From Fig. \ref{fourplot}(d) we can also have a better understanding of the system's behavior without modeling the event transition patterns with Event2Graph. Both scores become very similar across times. This phenomenon indicates the system have no understanding about the current state of the time series and cannot make a reasonable guess on what will happen next. In this case, the anomaly are purely reflected by the value change, instead of jointly considering the context of the value change like Event2Graph. 

\begin{figure}[ht]%

\includegraphics[width=8.5cm, height=4.5cm]{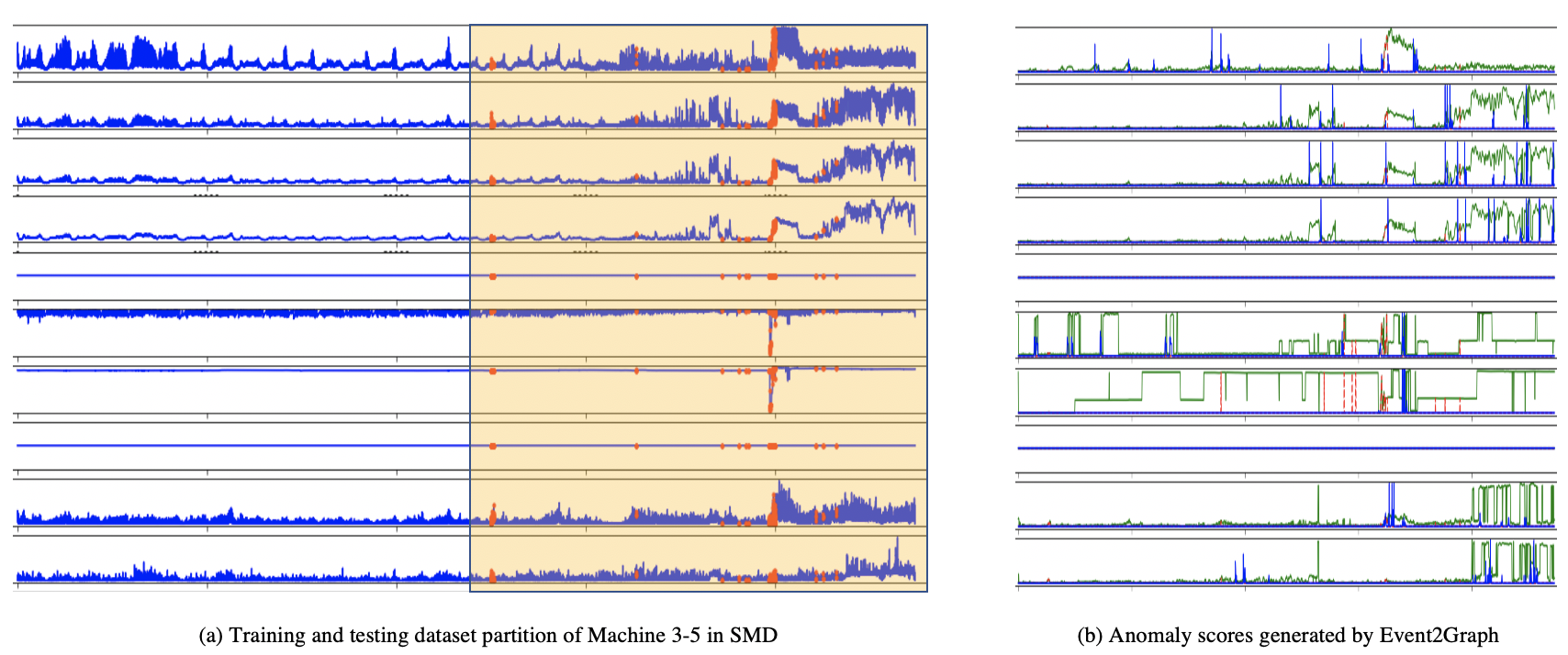}

\caption{Case study of a potential false negative. Testing region is highlighted in the left, and outputs of Event2Graph: $\omega_{1}$ (green line) and $\omega_{2}$ (blue line) are visualized on the right.}
\label{critic}
\end{figure}

Finally, we wish to highlight that our  Event2Graph system is able to capture the anomalies that have not been well marked by the ground-truth labels. For example in Fig. \ref{critic}, we visualize the anomaly scores generated by Event2Graph on a testing data in the SMD dataset. From the outputs of our system, it is clearly observed that there is an potential anomaly region after the ground-truth anomaly (red peak in Fig. \ref{critic}(a)). Our pattern matching score $\omega_{1}$ is significantly increased near the end of testing period, and also the residual error scores $\omega_{2}$ are frequently spiked. This phenomenon aligned well with human eye's inspection where there is an increasing, noisy trend emerging on three out of ten time series close to the end of testing period. We wish future works are able to re-label this dataset so that the ground-truth is able to reflect all types of the anomalies that corresponding to human's intuition.

\section{Conclusion}
We proposed an event-driven bipartite graph solution for multivariate time series anomaly detection. The solution does not assume any inter-dependency on time series, and all the relations are learned in a dynamic, data-driven manner. Our design is based on edge-stream so no adjacency matrix of the graph is required as input. As the system's memory is defined on the node level, our design left plenty space for future extensions such as inductive learning and parallel computation. Our solution achieved very competitive results on three anomaly detection datasets, and we encourage future works to explore further using bipartite event graph for multivariate anomaly detection.

\bibliographystyle{acm}
\bibliography{sample-base}

\end{document}